\documentclass[runningheads]{llncs}
\usepackage[T1]{fontenc}
\usepackage{graphicx}

\usepackage{cite}
\usepackage{amsmath,amssymb,amsfonts}
\usepackage{algorithmic}
\usepackage{graphicx}
\usepackage{textcomp}
\usepackage{xcolor}
\usepackage{subfigure}
\usepackage{multirow}
\begin{document}
\title{Gradient-Based Meta-Learning Using Uncertainty to Weigh Loss for Few-Shot Learning}
\titlerunning{Meta-Learning Using Uncertainty to Weigh Loss}
%
\author{Lin Ding\inst{1} \and
Peng Liu\inst{2} \and
Wenfeng Shen\inst{3} \and
Weijia Lu\inst{4} \and
Shengbo Chen\inst{5}}
\authorrunning{L. Author et al.}
%
\institute{Shanghai University, School of Computer Engineering and Science, Shanghai, China\\ 
\email{auroralin@shu.edu.cn} \and
United Automotive Electronic Systems, AI, Shanghai, China\\
\email{peng.liu2@uaes.com}\and
Xian Innovation College of Yanan University, Shanghai, China\\
\email{wfshen@shu.edu.cn}\and
United Automotive Electronic Systems, AI, Shanghai, China\\
\email{weijia.lu@uaes.com}\and
Shanghai University, School of Computer Engineering and Science, Shanghai, China\\ 
\email{schen@shu.edu.cn}}
\maketitle              
\begin{abstract}
Model-Agnostic Meta-Learning (MAML) is one of the most successful meta-learning techniques for few-shot learning. 
It uses gradient descent to learn commonalities between various tasks, enabling the model to learn the meta-initialization of its own parameters to quickly adapt to new tasks using a small amount of labeled training data. 
A key challenge to few-shot learning is task uncertainty. 
Although a strong prior can be obtained from meta-learning with a large number of tasks, a precision model of the new task cannot be guaranteed because the volume of the training dataset is normally too small. 
In this study, first, in the process of choosing initialization parameters, the new method is proposed for task-specific learner adaptively learn to select initialization parameters that minimize the loss of new tasks. 
Then, we propose two improved methods for the meta-loss part: Method 1 generates weights by comparing meta-loss differences to improve the accuracy when there are few classes, and Method 2 introduces the homoscedastic uncertainty of each task to weigh multiple losses based on the original gradient descent, as a way to enhance the generalization ability to novel classes while ensuring accuracy improvement. 
Compared with previous gradient-based meta-learning methods, our model achieves better performance in regression tasks and few-shot classification and improves the robustness of the model to the learning rate and query sets in the meta-test set.

\end{abstract}
\section{Introduction}
While recent deep learning methods have achieved better performance than humans on a variety of tasks, including image classification~\cite{1} or the game domain~\cite{2}, leveraged from large amount of data and huge computing resources. 
In many interesting scenarios, deep learning methods maynot achieve good results owing to the lack of large amounts of data~\cite{3}. 
Thus, meta-learning methods~\cite{4} have emerged as a potential solution to this problem by using information gathered from previous learning experiences to learn more effectively in a new task. 
This series of studies typically regards learning as a two-level process, each of which is aimed at a different scope. 
The meta-learner operates on the task level and gathers information from multiple task instances of a task-specific learner. 
Meanwhile, task-specific learners operate on the sample data level and incorporate the knowledge of the meta-learner into the learning process.

Model-Agnostic Meta-Learning (MAML)~\cite{5} is a classical meta-learning method that directly optimizes the gradient descent process of task-specific learners. 
All task-specific learners of MAML share initial parameters, and the meta-learner optimizes these initial parameters to perform fast gradient descent from these initial parameters to achieve good performance. 
However, this applies only if the sets of tasks are very similar. 
To solve this problem, we enable the task-specific learner to choose initialization parameters that minimize the loss of new tasks.

Simultaneous meta-learner learns from multiple tasks using a simple weighted sum of losses, where the loss weights are consistent or manually adjusted~\cite{6}. 
However, the performance of the meta-learner is highly dependent on the appropriate weight assignment to the loss of each task~\cite{7}. 
Determining the optimal weight is expensive and difficult to solve by manual adjustment. 
Therefore, we introduce an approach that combines multiple losses to simultaneously learn multiple tasks using homoscedastic uncertainty. 
We interpret homoscedastic uncertainty as task-dependent weights that can learn to balance various regression and classification losses. 
In contrast to learning each task in a balanced manner, our approach learns to balance these weights optimally, resulting in a superior performance.

The main contributions of this study are:
1. We enable task-specific learners to learn to select initialization parameters that minimize the loss of new tasks to merge task-specific information to better learn the commonalities among various tasks. 
2. Propose a weight generator by comparing meta-loss difference, showing the importance of loss weighting in gradient-based meta-learning. 
3. Leverage task homoscedastic uncertainty to learn and balance various losses for better performance.

\section{Related Work}
Many scholars have studied the problem of few-shot learning from the perspective of meta-learning~\cite{8}. 
Meta-learning provides a method to learn from a distribution of tasks and adapt to new tasks from the same distribution using a small number of samples~\cite{9}. 
There are four common categories of meta-learning methods:
1. Optimization-based meta-learning~\cite{10} provides learnable learning rules and optimization algorithms. 
2. Metric-based meta-learning~\cite{11} mainly learns task-related knowledge in the metric space. 
3. Model-based meta-learning~\cite{12} mainly identifies the task identity from several data samples and adapts to the task by adjusting the state of the model. 
4. Gradient-based meta-learning~\cite{13} focuses on learning model initialization such that new tasks can be quickly learned within a few gradient steps during testing.

Our method belongs to the fourth category above, but most gradient-based meta-learning algorithms~\cite{14} assume a globally shared initialization of all tasks, which leads to a key problem, namely, task uncertainty. 
Later, to explain uncertainty, researchers started to investigate the use of probabilistic models to customize a global shared initialization for each task~\cite{15} and merge task-specific information~\cite{16}. 
Similarly, our work also takes this into account so that the task-specific learner can choose the initialization parameters that minimize the loss of new tasks, which can make task-specific learners more sensitive to changes within different tasks. 
However, our inference uncertainty is different from the previous research direction of using probabilistic models~\cite{17}. 
Our method primarily considers the uncertainty of quantifying task-specific weights,  and introduces a meta-loss function with homoscedastic uncertainty.

\section{Background}
Our goal was to find a method to train a model that can quickly adapt to new tasks using a small amount of data and training iterations. 
As a cutting-edge technology for this purpose, model-agnostic meta-learning (MAML) is proposed by Finn et al.~\cite{5} and its details is list as following for the convenience of further discussion:

\vspace{-1em}
\subsection{Meta-learning problem setting}
In the few-shot learning problem, training tasks can be used to learn how to quickly adapt to a task with little data and then evaluate the (previously invisible) test task. 
The key assumption of the meta-learning algorithm is the need for a set of meta-tasks that share a common structure extracted from the distribution $\mathcal{P}\left(\mathcal{T} \right)$, so that they can be used to learn new tasks faster. 
During the training phase of the meta-learner, we obtained a set of tasks $\left\{ \mathcal{T}_{1},\mathcal{T}_{2},\mathcal{T}_{3}\ldots \right\}$. 
Each task $\mathcal{T}_{i}$ provided a support set $\left\{ \mathcal{D}_{\mathcal{T}_{i},train} \right\}$ for training the model and a query set $\left\{ \mathcal{D}_{\mathcal{T}_{i},test} \right\}$ to measure whether the training was effective (known as few-shot generalizability in the meta-training process), where the K-shot learning setup was used, assuming K samples for each class in the support and query sets.

In the meta-training process, sample a specific task $\mathcal{T}_{i} \in \left\{ \mathcal{T}_{1},~\mathcal{T}_{2},~\mathcal{T}_{3}\ldots \right\}$ from the distribution of task $\mathcal{P}\left(\mathcal{T} \right)$, train the model with K samples in the support set $\left\{ \mathcal{D}_{\mathcal{T}_{i},train} \right\}$, and use the loss $\mathcal{L}_{\mathcal{T}_{i}}\left( {\theta_{\mathcal{T}_{i}},\mathcal{D}_{\mathcal{T}_{i},train}} \right)$ corresponding to $\mathcal{T}_{i}$ as feedback to initialize the parameter $\theta_{\mathcal{T}_{i}}$ of the specific task model $f_{\theta_{\mathcal{T}_{i}}}$ (known as feedforward neural network in meta-learning setting), where the parameters obtained by optimizing $\mathcal{L}_{\mathcal{T}_{i}}\left( {\theta_{\mathcal{T}_{i}},\mathcal{D}_{\mathcal{T}_{i},train}} \right)$ are denoted by $\tilde{\theta}_{\mathcal{T}_{i}}$. 
Then, the feedback of the loss set on each query set $\left\{\mathcal{L}_{\mathcal{T}_{i}}\left(\tilde{\theta}_{\mathcal{T}_{i}}, \mathcal{D}_{\mathcal{T}_{i}, \text { test }}\right)\right\}_{\mathcal{T}_{i} \sim p(\mathcal{T})}$ is used to update the meta-learner $f_{\theta}$. 
At the end of meta-training, draw K samples in the new test task  $\mathcal{T}_{new}$ from $\mathcal{P}\left( \mathcal{T} \right)$, which helps the model $f_{\theta_{\mathcal{T}_{new}}}$ quickly adapt to $\mathcal{T}_{new}$ by gradient updating to measure the meta-performance of the model.

\vspace{-1em}
\subsection{Model-Agnostic Meta-Learning}
In tackling meta-learning problem, MAML adjusts a set of initial parameters $\theta$ using gradient-based learning rules to minimize network loss after several steps of gradient descent.
Specifically, the meta-learning objectives of MAML are as follows:
\begin{equation}
\vspace{-1.5em}
    \begin{split}
        \underset{\theta}{min}\,{\sum\limits_{\mathcal{T}_{i} \sim p(\mathcal{T})}\,}\mathcal{L}\left( {\theta - \alpha\nabla_{\theta}\mathcal{L}\left( {\theta,\mathcal{D}_{\mathcal{T}_{i},train}^{}} \right),\mathcal{D}_{\mathcal{T}_{i},test}^{}} \right) = \underset{\theta}{min}\,{\sum\limits_{\mathcal{T}_{i} \sim p(\mathcal{T})}\,}\mathcal{L}\left( {{\overset{\sim}{\theta}}_{\mathcal{T}_{i}},\mathcal{D}_{\mathcal{T}_{i},test}^{}} \right)\label{eq}
    \end{split}
\vspace{-1.5em}
\end{equation}

where ${\overset{\sim}{\theta}}_{\mathcal{T}i}$ denotes the parameters updated by the gradient descent. 
In particular, in the case of regression, the mean-squared error (MSE) loss function was used, and in the case of supervised classification, the cross-entropy loss function was used.

Let us consider a model $f_{\theta}$ parameterized by $\theta$.
Task-specific updates are performed in the inner loop.
When a new task $\mathcal{T}_{i}$ is given, the task-specific learner in MAML performs a gradient update of its parameters using the losses evaluated by the support set $\left\{ \mathcal{D}_{\mathcal{T}_{i},train} \right\}$, at which point the parameters of the model $\theta$ become task-specific parameters ${\overset{\sim}{\theta}}_{\mathcal{T}i}$.
\begin{equation}
    \left. {\overset{\sim}{\theta}}_{\mathcal{T}_{i}}\leftarrow\theta - \alpha\nabla_{\theta}\mathcal{L}_{\mathcal{T}}\left( {\theta,\mathcal{D}_{\mathcal{T}_{i},train}} \right) \right.
\end{equation}

The outer loop performs a meta-update of the cross-task meta-learner, and the model parameters $\theta$ are updated using the loss evaluated by the query set $\left\{ \mathcal{D}_{\mathcal{T}_{i},test} \right\}$.
\begin{equation}
\vspace{-0.2em}
    \left. \theta\leftarrow\theta - \beta\nabla_{\theta}\left( {{\sum\limits_{\mathcal{T} \sim p{(\mathcal{T})}}\,}\mathcal{L}_{\mathcal{T}}\left( {{\overset{\sim}{\theta}}_{\mathcal{T}_{i}},\mathcal{D}_{\mathcal{T}_{i},test}} \right)} \right) \right.
\end{equation}

where the sum of the losses is calculated using a small batch of tasks sampled from $\mathcal{P}\left( \mathcal{T} \right)$.

The final result of training is a well-learned initial parameter $\theta$ that is close to the local optimum of each task $\mathcal{T}_{i} \in \mathcal{P}\left( \mathcal{T} \right)$, so that any new task extracted from $\mathcal{P}\left( \mathcal{T} \right)$ can be adapted quickly in only a few gradient steps. 
Through this manner, the ${\overset{\sim}{\theta}}_{\mathcal{T}_{i}}$ obtained by the inner loop update would be sensitive to task-specific identification, because ${\overset{\sim}{\theta}}_{\mathcal{T}_{1}}$ and ${\overset{\sim}{\theta}}_{\mathcal{T}_{2}}$ exhibit different behaviors for different tasks $\mathcal{T}_{1},\mathcal{T}_{2} \in \mathcal{P}\left( \mathcal{T} \right)$.

\section{Meta-learning method}
In this section, we present our gradient-based meta-learning method using uncertainty to weigh loss, which can train the model to achieve rapid adaptation and handle the uncertainty that occurs when learning from small amounts of data, while being robust to changes in the learning rate and query sets in the meta-test set.

\vspace{-1em}
\subsection{Consolidate task-specific information}
MAML provides the same initialization for all tasks. 
However, this applies only if the task sets are very similar and task-specific information is ignored. 
To address this problem, we made task-specific learners learn to choose initialization parameters that minimize the loss of new tasks.

For MAML, the parameter search space of all tasks was fixed after the initial weights and network structures were determined. 
Therefore, under this condition, the parameters corresponding to all tasks can only be those of the model generated by the previous iteration, which limits the selection range of the parameters. 
This study proposes the addition of the selection action of task-related initialization parameters to the original MAML pipeline. 
First, we maintain a set $\left\{ \theta \right\}$ to save the parameters after meta-updating the gradient in the outer loop and then save the parameters after each iteration into this set, which is related to the tasks sampled in each epoch. 
Thus, $\theta_{0}$ is selected from this set $\left\{ \theta \right\}$ that minimizes $\mathcal{L}_{\mathcal{T}_{new}}\left( {\theta_{0},\mathcal{D}_{\mathcal{T}_{new}}} \right)$ as the initialization parameters for the next iteration, which is the parameter that makes the new task perform best. 
It can be seen that the selection range of parameters is projected into a task-related subspace, which makes the task-specific learner more sensitive to changes in the task and also enables our method to be robust to learning rates.

\vspace{-1em}
\subsection{Weight generator based on contrast meta-loss}
In MAML, we can see that its meta-learner simply resorts to a simple weight-consistent summation of losses during gradient descent, which is sensitive to the deviation of tasks and might inevitably degenerate over iterations. 
Then for a specific task , the observation can be classified into following categories: 
(1) high loss on the support set and high loss on the query set , (2) low loss on the support set but high loss on the query set, and (3) low loss on the support set and low loss on the query set. 
For the above three different situations, using only the query set for a simple summation of losses results in poor generalization. 
Therefore, we propose a weight generator based on contrast meta-loss, which focuses on comparing the loss on the support set and the loss on the query set for a specific task and using the ratio of the difference between the two losses to assign weights.

\begin{figure}
\vspace{-1em}
\includegraphics[width=\textwidth]{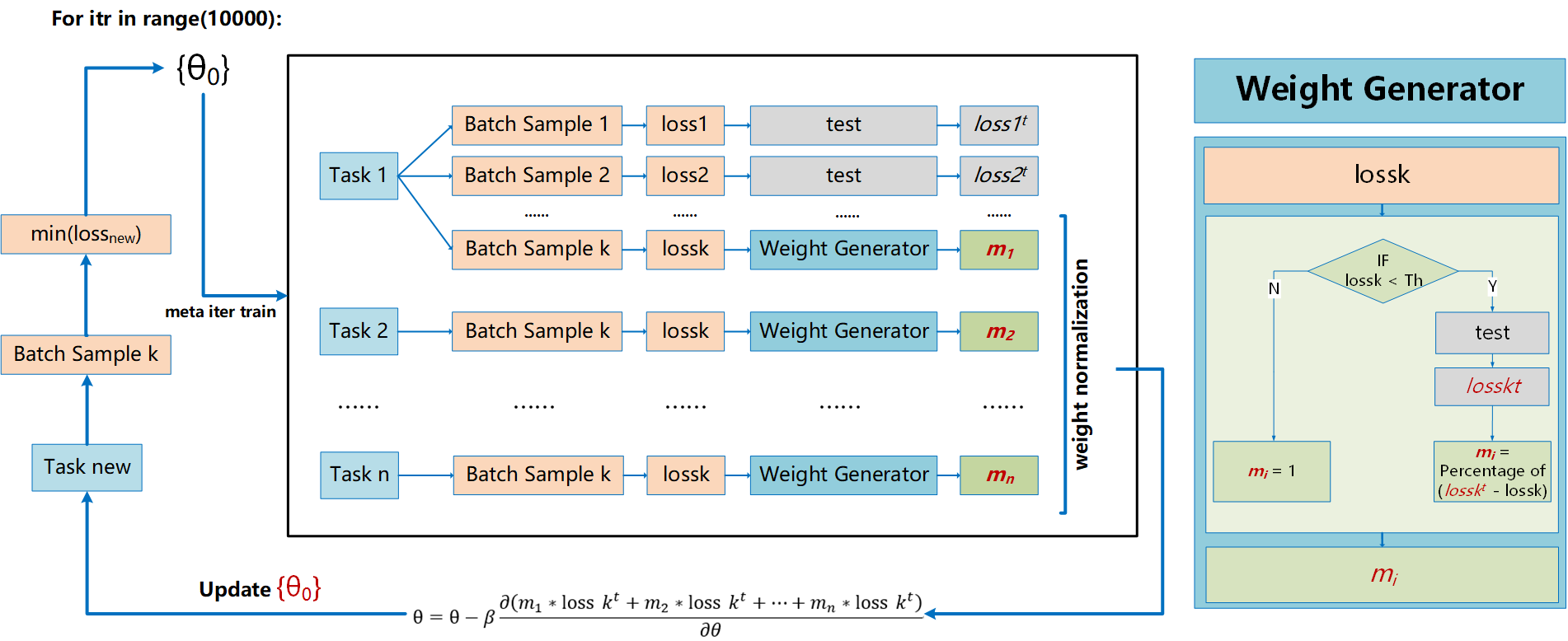}
\caption{Flowchart. We strengthen the importance of ${lossk}_{i}^{t}$ in the overall algorithm by adding a weight generator to produce significant optimization for the stability of gradient descent.} \label{fig4.1}
\vspace{-1em}
\end{figure}

Fig.~\ref{fig4.1} illustrates the flow of the proposed method. 
After calculating the meta-loss for each task, the subsequent steps were carried out. 
First, a threshold is given to the loss on the task support set to determine whether the task is trained well under a model with a certain initialization parameter. 
If it is higher than a given threshold, the training effect of the task is not good. 
The weight is directly assigned to one to retain the impact of the task on the overall loss so that the subsequent gradient can be decreased faster. 
If it is lower than the given threshold, it indicates that the task is better trained, and then the loss of the task on the query set is compared. 
At this point, weights can be assigned using the ratio of the loss differences. 
The larger the difference, the greater the weight, which causes the gradient of the task with poor generalization to descend faster and can ensure that the optimal point is reached after several steps of gradient descent. 
Finally, all weights are normalized to perform the meta-updating of the outer-loop cross-task meta-learner.

The weights are calculated by the following formula:
\begin{equation}
\vspace{-0.4em}
    m_{i} = \frac{{lossk}_{i}^{t} - {lossk}_{i}}{\sum_{i}^{n}{{lossk}_{i}^{t} - {lossk}_{i}}}
\end{equation}

where ${lossk}_{i}$ is the loss evaluated by the task-specific $Task~i$ using the kth sampled support set, and ${lossk}_{i}^{t}$ is the loss evaluated by $Task~i$ using the query set again after using ${lossk}_{i}$ to update the parameters with a gradient.

We show the meta-learning model with weight generator in Algorithm 1.

\begin{table}
\vspace{-1.5em}
\begin{center}
\begin{tabular}{l}
\hline\noalign{\smallskip}
\pmb{Algorithm 1} Meta-Learning Model with Weight Generator \\
\noalign{\smallskip}\hline\noalign{\smallskip}
Require: $\mathcal{P}\left( \mathcal{T} \right)$: distribution over tasks \\
Require: $\alpha,\beta$: step size hyperparameters \\
1: randomly initialize $\theta$ \\
2: while not done do \\
3: \ \ Sample batch of tasks $\mathcal{T}_{i} \in \mathcal{P}\left( \mathcal{T} \right)$ \\
4: \ \ for all $\mathcal{T}_{i}$ do \\
5: \ \ \ \ Evaluate $\nabla_{\{\theta_{0}\}}\mathcal{L}_{\mathcal{T}_{i}}\left( f\left( \theta \right) \right)$ with respect to K examples \\ 
\begin{tabular}[c]{@{}l@{}}
6: \ \ \ \ Compute adapted parameters with gradient descent: \\
\ \ \ \ \ \ \ \ $\theta_{i}^{'} = \theta - \alpha\nabla_{\{\theta_{0}\}}\mathcal{L}_{\mathcal{T}_{i}}\left( f\left( \theta \right) \right)$\end{tabular} \\
7: \ \ end for \\
8: \ \ Calculate the Proportion of loss difference $m_{i}$ and normalized \\
9: \ \ Update $\left. \theta\leftarrow\theta - \beta\nabla_{\{\theta_{0}\}}\left( {{\sum\limits_{\mathcal{T}_{i} \sim \mathcal{P}{(\mathcal{T})}}\,}{m_{i}^{}\mathcal{L}}_{\mathcal{T}_{i}}\left( {f\left( \theta_{i}^{'} \right)} \right)} \right) \right.$ \\
\begin{tabular}[c]{@{}l@{}}
10: \ Store $\theta$ into the collection $\left\{ \theta_{0} \right\}$ and choose the best $\theta_{0}$ \\
\ \ \ \ \ \ with the least $\mathcal{L}_{\mathcal{T}_{new}}\left( {\theta_{0},\mathcal{D}_{\mathcal{T}_{new}}} \right)$\end{tabular} \\
11: end while \\
\noalign{\smallskip}\hline
\end{tabular}
\end{center}
\vspace{-1.5em}
\end{table}

\vspace{-1.5em}
\subsection{Gradient descent based meta-learning with homoscedastic uncertainty}
The use of few-shot learning inevitably produces models with high uncertainties. 
Consequently, the learned model may not be able to make high-confidence predictions for new tasks. 
In this section, we introduce a meta-loss function based on Gaussian likelihood maximization with homoscedastic uncertainty.

Homoscedastic uncertainty~\cite{7} is independent of the output of the model, captures noise in the task, and is a quantity that remains constant for all input data but varies across tasks. 
Therefore, it can be described as a task-related uncertainty. 
In the meta-setting, homoscedastic uncertainty is caused by task-related weights, from which we can use homoscedastic uncertainty as a basis for weighting losses in meta-learning problems. 
In the meta-loss function, we used inter-task homoscedasticity as the observed inter-task noise to capture the relative confidence between tasks, reflecting the inherent uncertainty of regression or classification tasks.

First, we define $\mathbf{f}^{w}\left( \mathbf{x} \right)$ as the output of a network with input x weighted by $w$. 
For classification, the model output is compressed by a Softmax function and sampled from the resulting probability vector:
\begin{equation}
\vspace{-0.4em}
    p\left( {\mathbf{y} \mid \mathbf{f}^{w}\left( \mathbf{x} \right),\sigma} \right) = {\rm Softmax}\left( {{\frac{1}{\sigma^{2}}}\mathbf{f}^{w}\left( \mathbf{x} \right)} \right)
\end{equation}

Where the inputs are scaled by $\sigma^{2}$ (often also referred to as temperature scaling), which is related to the uncertainty of the task measured in terms of entropy. 
At this point, we define the log-likelihood of the model output as
\begin{equation}
\vspace{-0.4em}
\begin{split}
	{\rm log}p\left( {\mathbf{y} = c \mid \mathbf{f}^{w}\left( \mathbf{x} \right),\sigma} \right) &= \frac{1}{\sigma^{2}}\mathbf{f}_{c}^{w}\left( x \right) - {\rm log}{\sum\limits_{c^{'}}\,}{\rm exp}\left( {\frac{1}{\sigma^{2}}\mathbf{f}_{c^{'}}^{w}\left( \mathbf{x} \right)} \right)
\end{split}
\end{equation}

Where $\mathbf{f}_{c}^{w}\left( \mathbf{x} \right)$ denotes the cth element of the vector $\mathbf{f}^{w}\left( \mathbf{x} \right)$.

Next, our model output is composed of multiple classification task vectors $y_{1},\ldots,y_{n}$, which are modeled by the softmax likelihood.
This leads to the minimization objective of our model, $\mathcal{L}\left( {w,\sigma_{1},\ldots,\sigma_{n}} \right)$ (classification loss):
\begin{equation}
\vspace{-0.4em}
    \begin{split}
        &= - {\rm log}p\left( {\mathbf{y}_{1} = c_{1},\ldots , \mathbf{y}_{n} = c_{n} \mid \mathbf{f}^{w}\left( \mathbf{x} \right)} \right)\\
        &= {\rm Softmax}\!\left( {\mathbf{y}_{1} \!=\! c_{1};\mathbf{f}^{w}\!\left( \!\mathbf{x}\! \right)\!,\sigma_{1}} \!\right)\!\cdot\! \cdot\! \cdot\! {\rm Softmax}\!\left( {\mathbf{y}_{n} \!=\! c_{n};\mathbf{f}^{w}\!\left(\! \mathbf{x} \!\right)\!,\sigma_{n}}\! \right)\!\\
        &= \frac{1}{\sigma_{1}^{2}}\mathcal{L}_{1}\left( w \right) + {\log{f\left( \sigma_{1} \right)}} + \cdot \cdot \cdot + \frac{1}{\sigma_{n}^{2}}\mathcal{L}_{n}\left( w \right) + {\log{f\left( \sigma_{n} \right)}}\\
        &\approx \frac{1}{\sigma_{1}^{2}}\mathcal{L}_{1}\left( w \right) + \cdot \cdot \cdot + \frac{1}{\sigma_{n}^{2}}\mathcal{L}_{n}\left( w \right) + \log\sigma_{1} + \cdot \cdot \cdot + \log\sigma_{n}\\
        &= {\sum\limits_{i}^{n}\left( \frac{1}{\sigma_{i}^{2}}\mathcal{L}_{i}\left( w \right) + {\rm log}\sigma_{i} \right)}\\
    \end{split}
\end{equation}

Here, we use $\mathcal{L}_{i}\left( w \right) = - \log {\rm Softmax}\left( {\mathbf{y}_{i},\mathbf{f}^{w}\left( \mathbf{x} \right)} \right)$ to denote the cross-entropy loss of $\mathbf{y}_{i}$.
In the equation, we let $f\left( \sigma_{i} \right) = \frac{{\sum\limits_{{c_{i}}^{'}}\,}{\exp\left( {\frac{1}{\sigma_{i}^{2}}f_{{c_{i}}^{'}}^{w}\left( \mathbf{x} \right)} \right)}}{\left( {{\sum\limits_{{c_{i}}^{'}}\,}{\exp\left( {f_{{c_{i}}^{'}}^{w}\left( \mathbf{x} \right)} \right)}} \right)^{\frac{1}{\sigma_{i}^{2}}}}$ and introduce a simplifying assumption $\frac{1}{\sigma_{i}}{\sum\limits_{{c_{i}}^{'}}\,}{\rm exp}\left( {\frac{1}{\sigma_{i}^{2}}\mathbf{f}_{{c_{i}}^{'}}^{w}\left( \mathbf{x} \right)} \right) \approx \left( {{\sum\limits_{{c_{i}}^{'}}\,}{\rm exp}\left( {\mathbf{f}_{{c_{i}}^{'}}^{w}\left( \mathbf{x} \right)} \right)} \right)^{\frac{1}{\sigma_{i}^{2}}}$, which becomes an equation when $\left. \sigma_{2}\rightarrow 1 \right.$.
This simplifies the optimization objectives and empirically improves the results. 
Where $\frac{1}{\sigma_{i}^{2}}$ is defined as the relative weight of the adaptive learning loss $\mathcal{L}_{i}\left( w \right)$ through the data, that is, the relative confidence between tasks, and the weight of $\mathcal{L}_{i}\left( w \right)$  decreases as $\sigma_{i}$ (observation noise scalar of variable $\mathbf{y}_{i}$) increases.
At the same time, the last item of the target $\sum_{i}^{n}{{\rm log}\sigma_{i}}$ is the regularizer of the noise scalar that is added to prevent the noise from increasing excessively, which can effectively ignore the impact of abnormal data and thus improve the robustness of the algorithm when the query set changes.

We show the meta-learning model with homoscedastic uncertainty in Algorithm 2.
\begin{table}
\vspace{-1.5em}
    \begin{center}
        \begin{tabular}{l}
            \hline\noalign{\smallskip}
            \pmb{Algorithm 2} Meta-Learning Model with Homoscedastic Uncertainty                                                                                                                                                                                                                     \\
            \noalign{\smallskip}\hline\noalign{\smallskip}
            Require: $\mathcal{P}\left( \mathcal{T} \right)$: distribution over tasks                                                                                                                                                                           \\
            Require: $\alpha,\beta$: step size hyperparameters                                                                                                                                                                                                  \\
            1: randomly initialize $\theta$                                                                                                                                                                                                                           \\
            2: while not done do                                                                                                                                                                                                                         \\
            3: \ \ Sample batch of tasks $\mathcal{T}_{i} \in \mathcal{P}\left( \mathcal{T} \right)$                                                                                                                                                                  \\
            4: \ \ for all $\mathcal{T}_{i}$ do               \\
            5: \ \ \ \ Evaluate $\nabla_{\{\theta_{0}\}}\mathcal{L}_{\mathcal{T}_{i}}\left( f\left( \theta \right) \right)$ with respect to K examples    \\
            \begin{tabular}[c]{@{}l@{}}
            6: \ \ \ \ Compute adapted parameters with gradient descent: \\
                \ \ \ \ \ \ \ \ $\theta_{i}^{'} = \theta - \alpha\nabla_{\{\theta_{0}\}}\mathcal{L}_{\mathcal{T}_{i}}\left( f\left( \theta \right) \right)$\end{tabular}                                                                                                                                                                                                                                 \\
            7: \ \ end for                                                                                                                                                                                                                                          \\
            8: \ \ Compute model weight $w$ and noise scalar $\sigma_{i}$ of $\mathcal{T}_{i}$   \\
            \ \ \ \ \ for maximum likelihood estimation ${\rm log}p\left( {\mathbf{y} \mid \mathbf{f}^{w}\left( \mathbf{x} \right),\sigma_{i}} \right)$  \\
            9: \ \ Compute the minimization objective of our model $\mathcal{L}\left( {w,\sigma_{1},\ldots,\sigma_{n}} \right)$ \\
            10: \ Update $\left. \theta\leftarrow\theta - \beta\nabla_{\{\theta_{0}\}}\mathcal{L}\left( {w,\sigma_{1},\ldots,\sigma_{n}} \right) \right.$ \\
            \begin{tabular}[c]{@{}l@{}}
            11: \ Store $\theta$ into the collection $\left\{ \theta_{0} \right\}$ and choose the best $\theta_{0}$ \\
                \ \ \ \ \ \ with the least $\mathcal{L}_{\mathcal{T}_{new}}\left( {\theta_{0},\mathcal{D}_{\mathcal{T}_{new}}} \right)$\end{tabular}                                                                                                                                                                                                                                 \\
            12: end while                                                                                                                                                                                                                                       \\
            \noalign{\smallskip}\hline
        \end{tabular}
    \end{center}
\vspace{-3.5em}
\end{table}

\section{Experiment}
We conducted experiments to verify:1. Our method can be used for meta-learning in several domains including supervised regression and classification. 2. Our method was robust to task-specific learning rates and the number of query sets in the meta-test set without overfitting. 3. Our method can improve the meta-learning performance.

Simple regression and classical few-shot image classification problems were studied in our experiments. For these experimental domains, the MAML and our meta-learning method were compared. Most of our experiments were completed by modifying the code attached to MAML~\cite{5}; therefore, both our meta-learning method and MAML used the same neural network structure and the same number of internal gradient steps.

\vspace{-1em}
\subsection{Regression experiments}
We begin with a simple regression problem that illustrates the basic principles of MAML.
Each task involves regressing from the input to the output of a sine wave, where the amplitude and phase of the sine wave vary from task to task.
Thus, $\mathcal{P}\left( \mathcal{T} \right)$ is continuous, in which the amplitude varies within [0.1, 5.0], the phase varies within [0, $\pi$], and the dimensions of both the input and output are 1.
During the training and testing, the data points are uniformly sampled from [-5.0, 5.0].
The loss is the mean square error between the predicted $f\left( x \right)$ and true value.
For training, we used a sine function of K = 10 data points with a fixed step size $\alpha$ = 0.01 as the regression task, and Adam as the meta-optimizer~\cite{18}.

\begin{figure}
\vspace{-1em}
  \centering
  \subfigure[MAML]{
  \begin{minipage}{0.47\linewidth}
  \centering
  \includegraphics[width=1\linewidth]{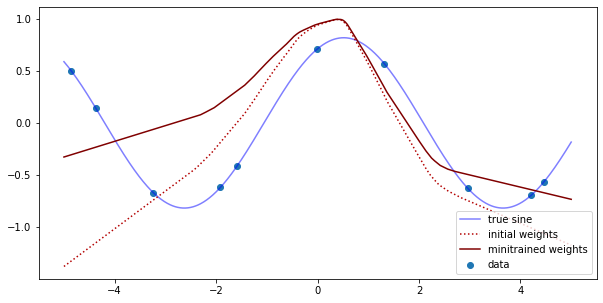}
  \end{minipage}
  }  
  \subfigure[Our model]{
  \begin{minipage}{0.47\linewidth}
  \centering
  \includegraphics[width=1\linewidth]{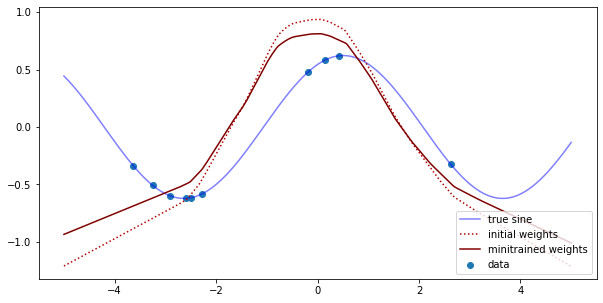}  
  \end{minipage} 
  }
  \vspace{-1em}
  \caption{Few-shot learning for a simple regression task. Meta-preprocessing of the network with 10 regression tasks.The red dotted line represents the curve of the model after fitting to 10 data points at the initial weights, and the solid red line represents the curve of the model after fitting 10 data points with weights trained with MAML(a) or Our model(b).It can be seen that our model (b) is closer to the true sine than MAML (a) after training.}
  \label{fig5.1}
\vspace{-2em}
\end{figure}

In the above regression task, we evaluated the performance using the model learned by the MAML and our model. During the fine-tuning process, K data points are randomly sampled to calculate each gradient step. The qualitative results are shown in Fig.~\ref{fig5.1}. Our model (b) is able to quickly adapt to the sinusoidal regression with only 10 data points after training, whereas the MAML-trained model (a) cannot adequately adapt to with very few data points after training.

\vspace{-1em}
\subsection{Classification experiments}
To compare the performance of our method with previous meta-learning and few-shot learning algorithms, we evaluated them on few-shot classification on the miniImageNet~\cite{19} and tieredImageNet~\cite{20} datasets.

Dataset: miniImageNet contains 100 semantic classes with 600 images per class, where all images are resized to 84×84. In the experiments, these 100 classes were divided into meta-training/validation/test sets with out of out of 64/16/20 (nonoverlapping) classes. That is, there were 64 base classes and 20 novel classes, and the other 16 classes were used for hyperparameter tuning. tieredImageNet is a larger subset of ILSVRC-2012~\cite{21}, containing 608 semantic classes in 34 superclasses with an image size of 84×84 in the classes. In the experiments, the 34 superclasses were divided into 20/6/8, corresponding to the meta-training/validation/test sets of 351/97/160 classes. Because the base classes and novel classes come from different superclasses, the classification method will be more challenging for this dataset.

Network architecture: We used a CNN classification model with the same architecture as the MAML. The model has four modules: each module has 3×3 convolutions and 32 filters (to reduce overfitting), followed by batch normalization~\cite{22}, ReLU nonlinear layers, and a 2×2 max pooling layer.

Training and evaluation: During meta-training, meta-testing, meta-validation, we sample 5-way 1-shot tasks from the corresponding classes and images. Every time we sampled five different classes, we randomly assigned index c$\in$[N] to each class. During meta-testing, we sampled 10,000 tasks. For each task, the query set contained 15 images per class. We report the results of our method on the miniImageNet and tieredImageNet datasets compared to previous work in Table~\ref{tab1} and Table~\ref{tab2}, including the average few-shot classification accuracy ($\%$) and 95$\%$ confidence interval.

\begin{table}
\vspace{-1.2em}
\caption{miniImageNet average few-shot classification accuracy}\label{tab1}
\begin{tabular}{l|c|c|c}
\hline
\multirow{2}{*}{Models} & \multirow{2}{*}{Backbone} & \multicolumn{2}{|c}{miniImageNet 5-way}\\
\cline{3-4}& & 1 shot& 5 shot\\
\hline
Fine-tune baseline &  ConvNet-4 & 28.86 ± 0.54$\%$ & 49.79 ± 0.79$\%$ \\
Matching Networks~\cite{9} &  ConvNet-4 & 43.56 ± 0.84$\%$ & 55.31 ± 0.73$\%$\\
Meta-Learning LSTM~\cite{10} & ConvNet-4 & 43.44 ± 0.77$\%$ & 60.60 ± 0.71$\%$\\
MAML~\cite{5} & ConvNet-4 & 48.70 ± 1.84$\%$ & 63.11 ± 0.92$\%$\\
Prototypical Networks~\cite{17} & ConvNet-4 & 49.42 ± 0.78$\%$ & \textbf{68.20 ± 0.66$\%$}\\
Relation Networks~\cite{11} & ConvNet-4 &50.44 ± 0.82$\%$  & 65.32 ± 0.70$\%$\\
R2D2~\cite{23} & ConvNet-4 & 51.20 ± 0.60$\%$ & \textbf{68.80 ± 0.10$\%$}\\
MetaOptNet~\cite{24} & ConvNet-4 & 52.87 ± 0.57$\%$ & \textbf{68.76 ± 0.48$\%$}\\
MAML+Meta-dropout~\cite{25} & ConvNet-4 & 51.93 ± 0.67$\%$ & \textbf{67.42 ± 0.52$\%$}\\
Negative Margin~\cite{26} & ConvNet-4 & 52.84 ± 0.76$\%$ & \textbf{70.41 ± 0.66$\%$}\\
\hline
Ours(Weight generator) & ConvNet-4 & 52.89 ± 1.82$\%$ & 65.95 ± 0.96$\%$\\
\textbf{Ours(Homoscedastic uncertainty)} & \textbf{ConvNet-4} & \textbf{53.97 ± 1.80$\%$} & \textbf{71.62 ± 0.43$\%$}\\
\hline
\end{tabular}
\vspace{-1.5em}
\end{table}

\begin{table}
\vspace{-2em}
\caption{tieredImageNet average few-shot classification accuracy}\label{tab2}
\begin{tabular}{l|c|c|c}
\hline
\multirow{2}{*}{Models} & \multirow{2}{*}{Backbone} & \multicolumn{2}{|c}{miniImageNet 5-way}\\
\cline{3-4}
 & & 1 shot& 5 shot\\
\hline
MAML~\cite{5} & ConvNet-4 & 51.67 ± 1.81$\%$ & 70.30 ± 1.75$\%$\\
Prototypical Networks~\cite{17} & ConvNet-4 & 53.31 ± 0.89$\%$ & \textbf{72.69 ± 0.74$\%$}\\
Relation Networks~\cite{11} & ConvNet-4 &\textbf{54.48 ± 0.93$\%$}  & 71.32 ± 0.78$\%$\\
	MetaOptNet~\cite{24} & ConvNet-4 & \textbf{54.71 ± 0.67$\%$} & 71.79 ± 0.59$\%$\\
\hline
Ours(Weight generator) & ConvNet-4 & 53.9 ± 1.89$\%$ & 72.22 ± 1.03$\%$\\
\textbf{Ours(Homoscedastic uncertainty)} & \textbf{ConvNet-4} & \textbf{55.37 ± 0.78$\%$} & \textbf{74.08 ± 1.65$\%$}\\
\hline
\end{tabular}
\vspace{-1.5em}
\end{table}

Our main idea is to improve the accuracy of the algorithm by changing the weights of the meta-losses; therefore, we first compare the difference between the meta-losses to generate the weights, and find that our (weight generator) method has a certain performance improvement, but it is still not as good as the current state-of-the-art algorithm for few-shot classification tasks. Therefore, we introduce homoscedastic uncertainty to the original idea. As can be seen from the tables, the accuracy of the model learned by our (homoscedastic uncertainty) method is significantly improved on both the miniImageNet and tieredImageNet datasets, with an overall improvement of 1–2$\%$ over the advanced algorithms mentioned above.

\begin{figure}
\vspace{-1.5em}
\centering
\subfigure{
\begin{minipage}{0.45\linewidth}
\centering
\includegraphics[width=1\linewidth]{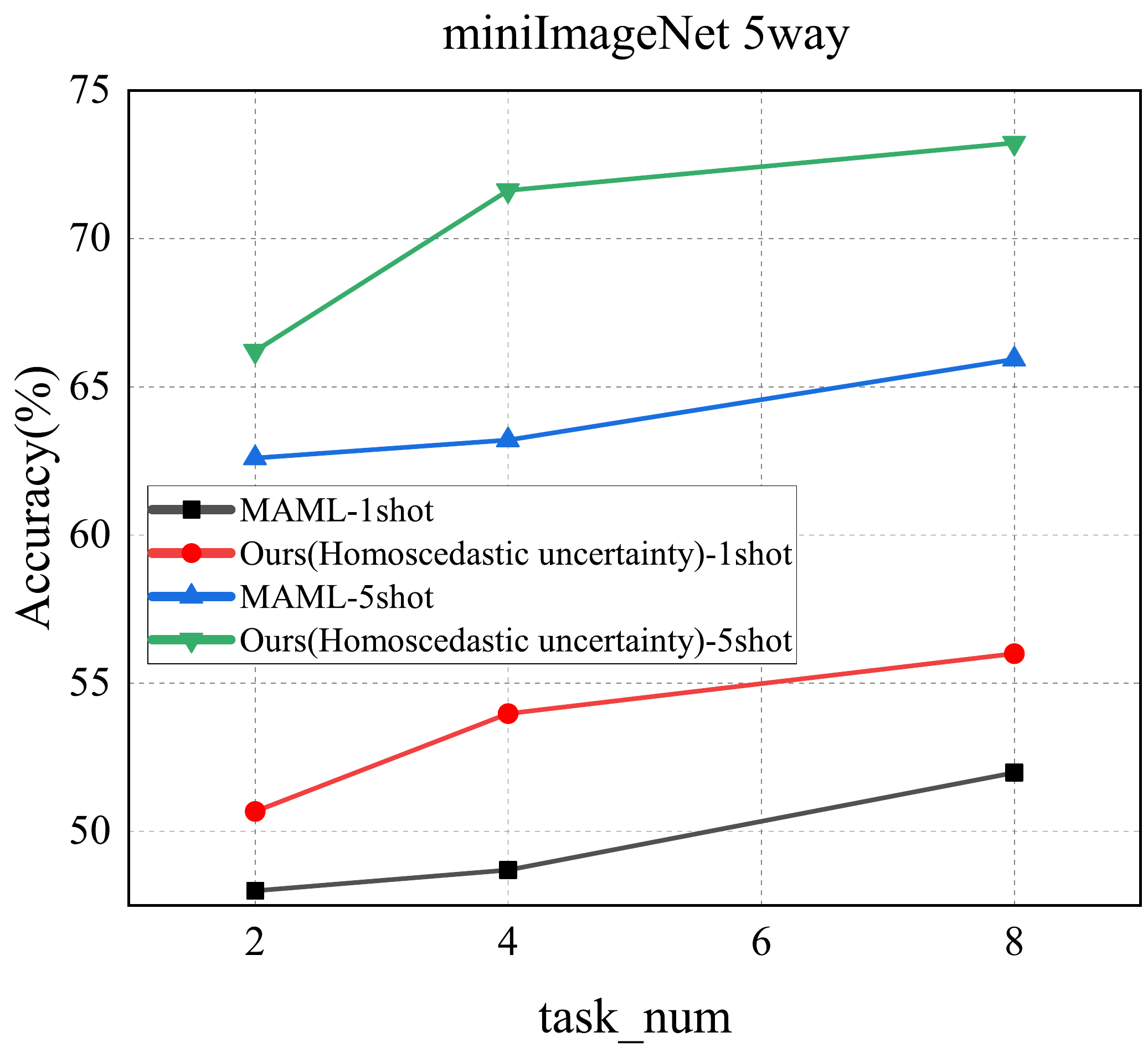}
\end{minipage}
} 
\subfigure{
\begin{minipage}{0.45\linewidth}
\centering
\includegraphics[width=1\linewidth]{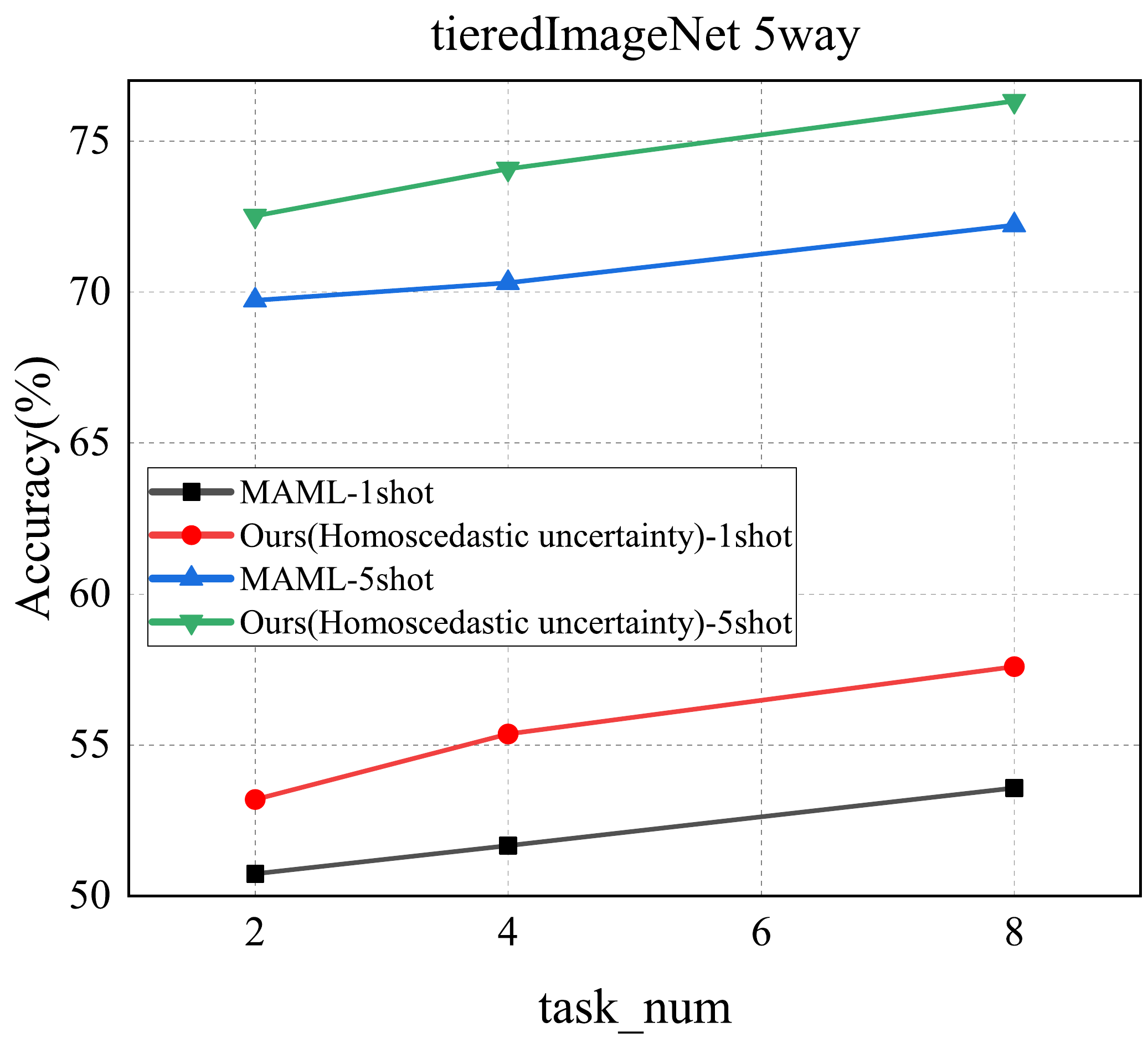} 
\end{minipage} 
}
\vspace{-1em}
\caption{Average few-shot classification accuracy ($\%$) for MAML and our (Homoscedastic uncertainty) method under different number of tasks. The training of our (Homoscedastic uncertainty) method on 4 or 8 tasks has a higher accuracy increase than its training on 2 tasks.}
\label{fig5.2}
\vspace{-3em}
\end{figure}

We also set different task numbers based on the above experiments for the next comparison. From Fig.~\ref{fig5.2}, we can see that the performance improvement of our (homoscedastic uncertainty) method is greater with a larger number of tasks (task$\_$num=4) because homoscedastic uncertainty can capture the noise between tasks as a representation of the weight of a specific task. When the number of tasks is large, the weight distribution is more balanced, the task information learned by meta-loss is greater, and the initialization parameters learned by meta-learning are better.

\vspace{-0.2em}
\subsection{Robustness to learning rate and query set changes}
Based on the regression and classification experiments in Sections 5.1 and 5.2, we set different step sizes ($\alpha \in \left\{ 10^{- 3},10^{- 2}, 10^{- 1} \right\}$) to perform multiple groups of comparative experiments to observe the adaptability of our method to different learning rates $\alpha$ compared to MAML.

\begin{figure}
\vspace{-1.5em}
  \centering
  \subfigure[]{
  \begin{minipage}{0.45\linewidth}
  \centering
  \includegraphics[width=1\linewidth]{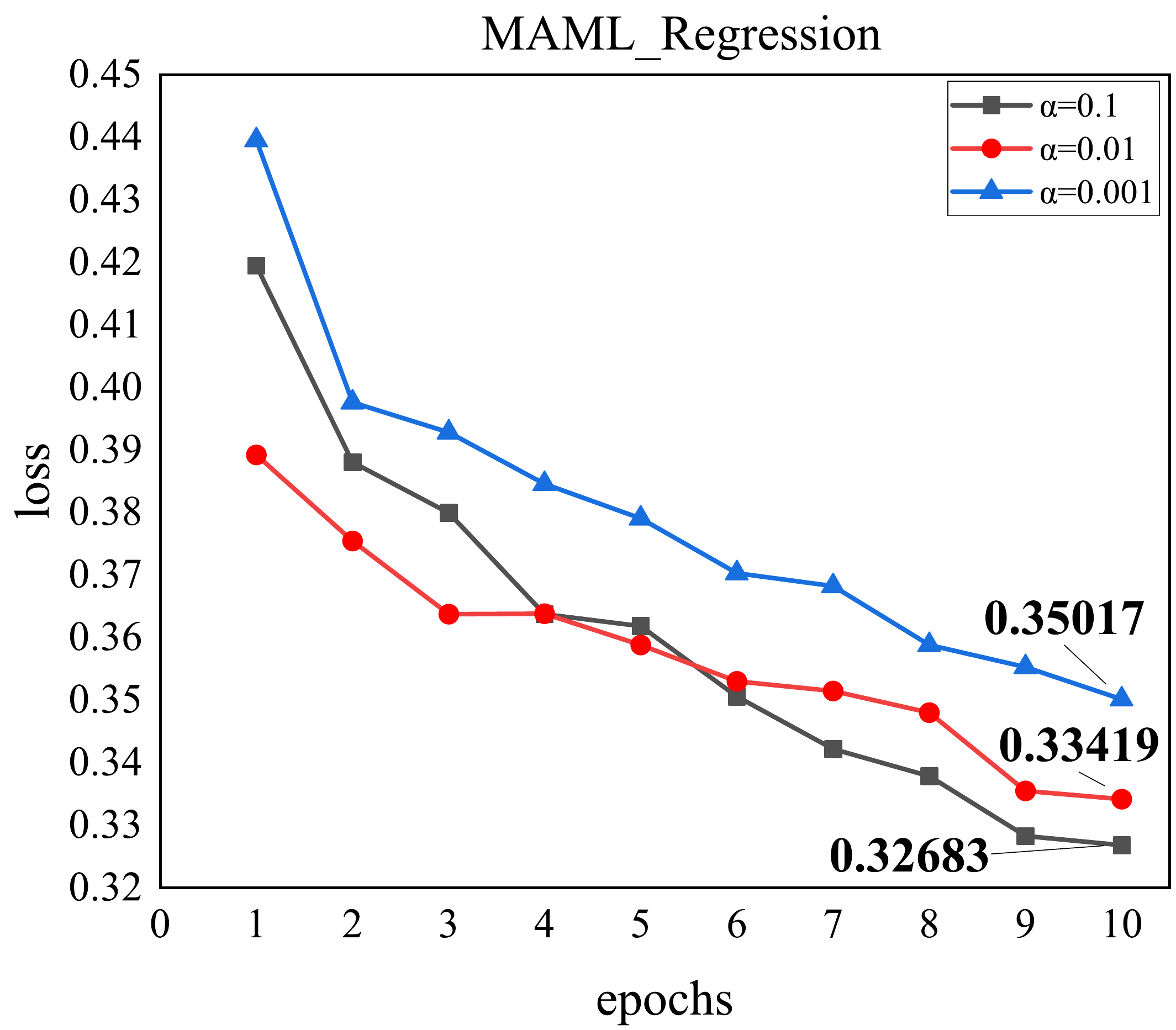}
  \end{minipage}
  }  
  \subfigure[]{
  \begin{minipage}{0.45\linewidth}
  \centering
  \includegraphics[width=1\linewidth]{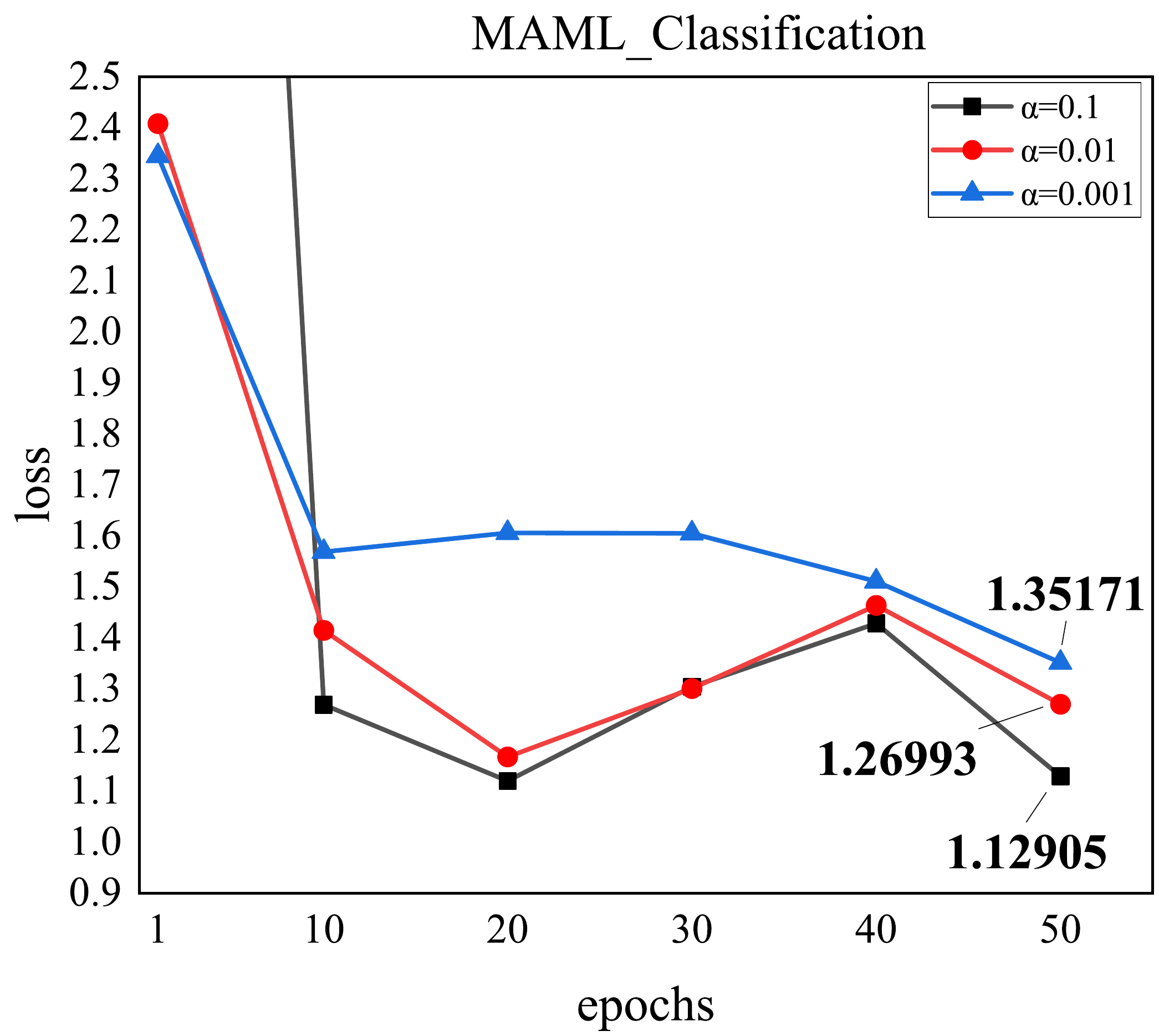}  
  \end{minipage} 
  }
  \subfigure[]{
  \begin{minipage}{0.45\linewidth}
  \centering
  \includegraphics[width=1\linewidth]{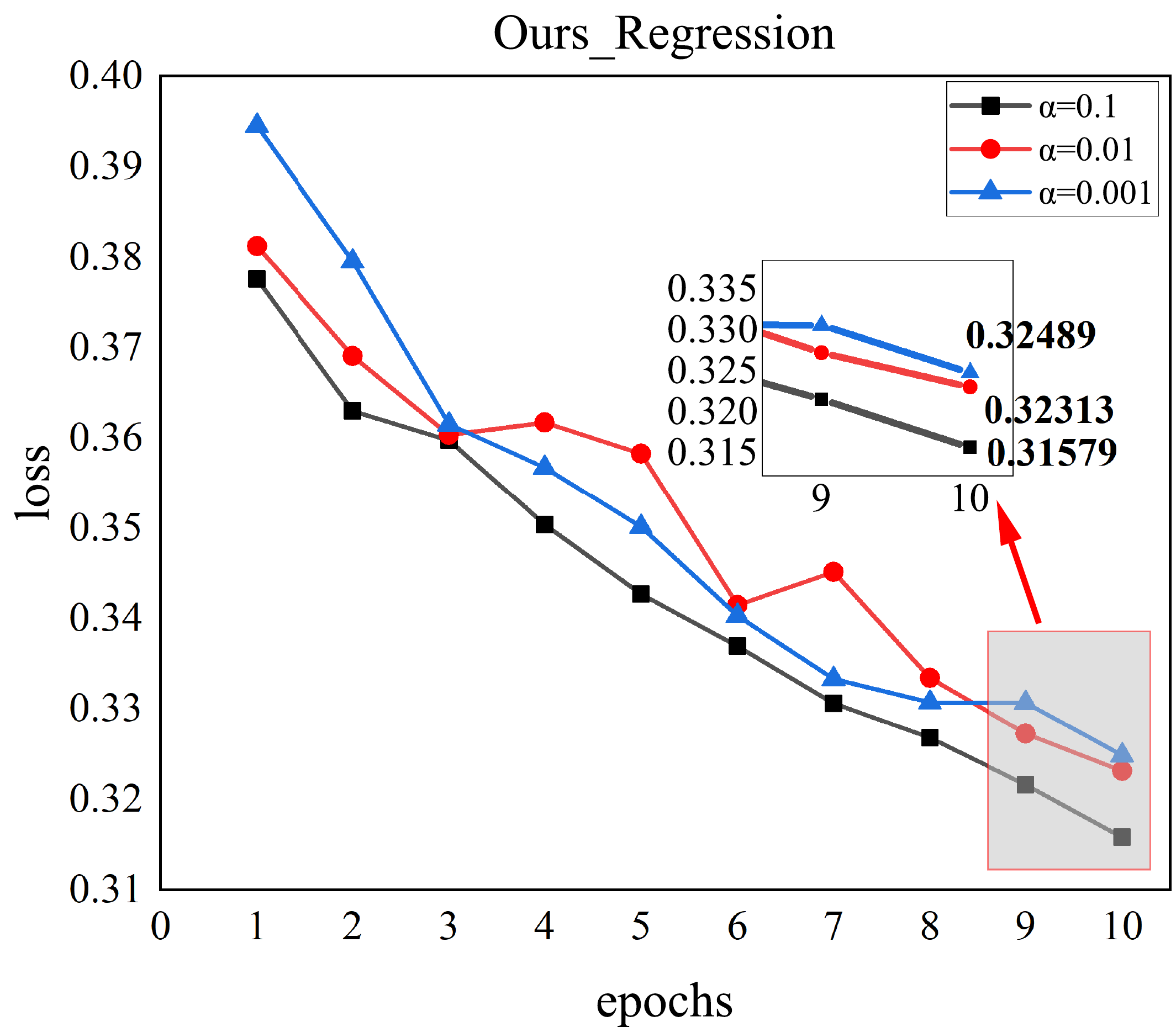}
  \end{minipage}
  }  
  \subfigure[]{
  \begin{minipage}{0.45\linewidth}
  \centering
  \includegraphics[width=1\linewidth]{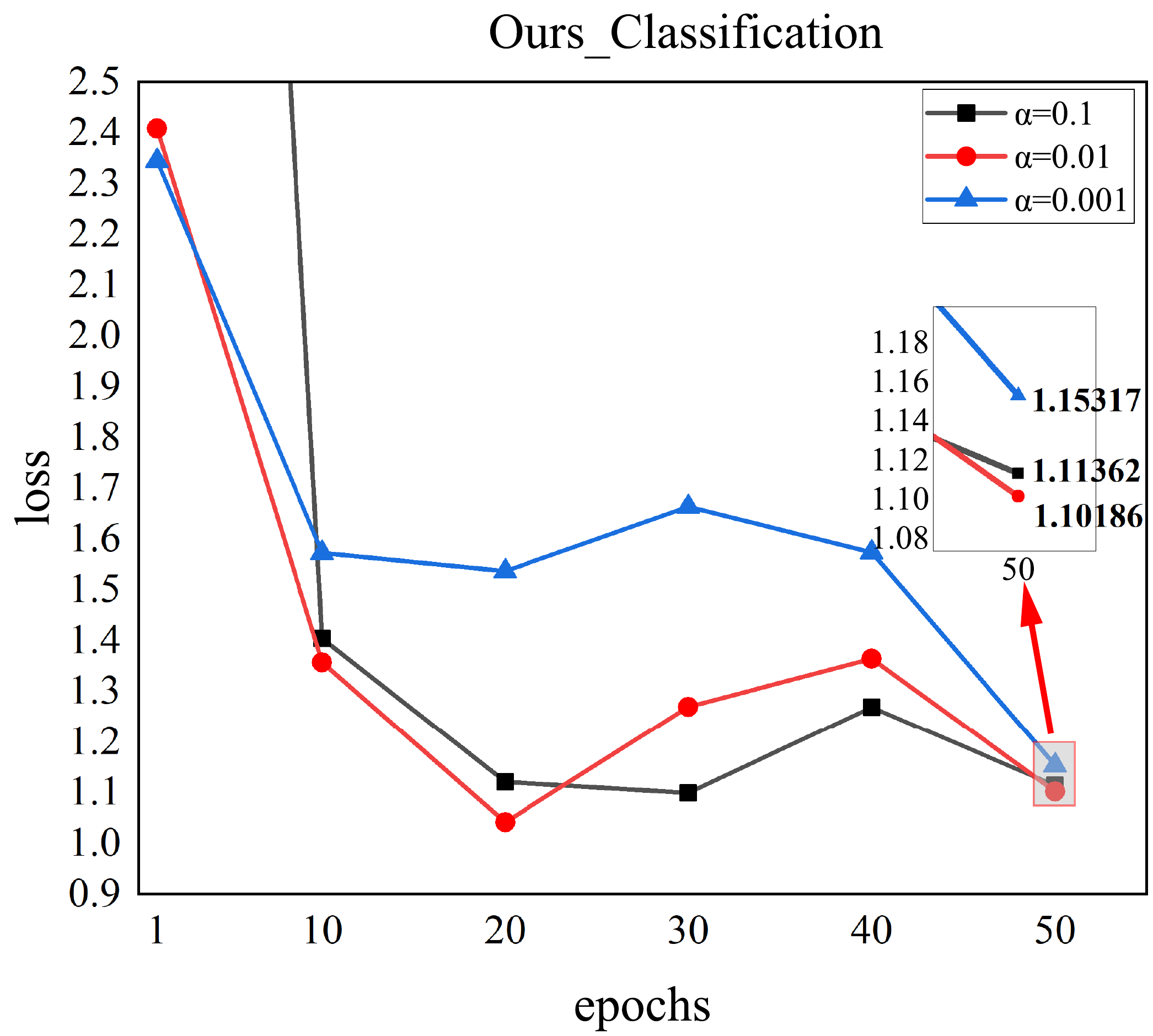}  
  \end{minipage} 
  }
  \vspace{-1em}
  \caption{Loss curves of MAML and our method for regression and classification: Loss at 10 sine wave regression (left), loss at 5-way 1-shot image classification (right).}
  \label{fig5.4}
\vspace{-3em}
\end{figure}

As shown in Fig.~\ref{fig5.4}, at 10 sine wave regressions, the losses of our method (c) at $\alpha = 10^{- 3}$ and $\alpha = 10^{- 2}$ are concentrated, but MAML (a) differs by 0.015 at $\alpha = 10^{- 3}$ versus $\alpha = 10^{- 2}$, and at $\alpha = 10^{- 1}$ our method can make the loss reach the lowest value of 0.315, which cannot be achieved by the MAML algorithm. In 5-way 1-shot image classification, our method (d) at $\alpha \in \left\{ 10^{- 3},10^{- 2}, 10^{- 1} \right\}$ respectively reaches a fixed value of the final loss compared to MAML (b), where the final loss is in a divergent state. Experiments demonstrate that our model is more robust to changes in learning rate $\alpha$.

\begin{figure}
\vspace{-1em}
  \centering
  \subfigure{
  \begin{minipage}{0.45\linewidth}
  \centering
  \includegraphics[width=1\linewidth]{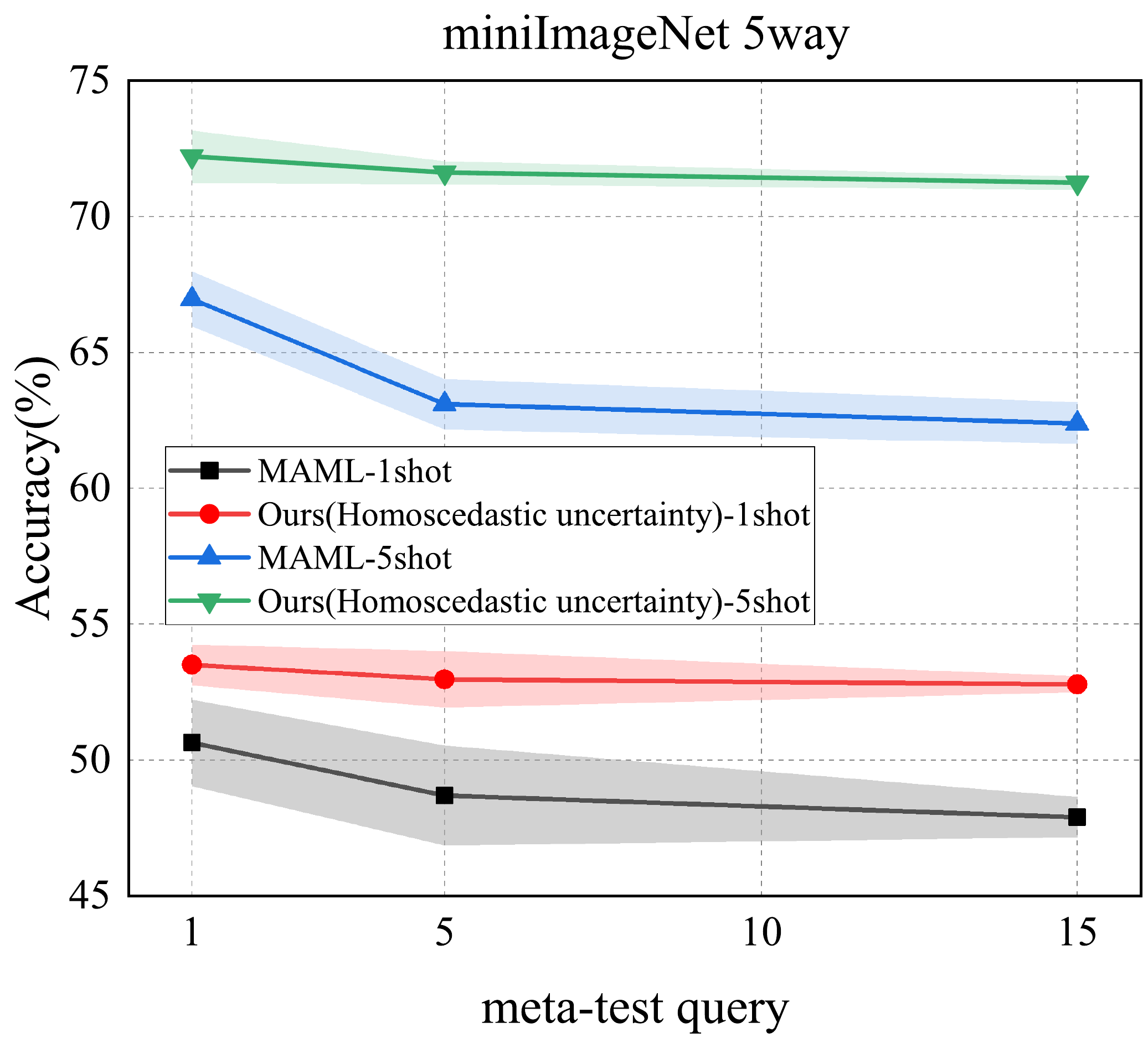}
  \end{minipage}
  }  
  \subfigure{
  \begin{minipage}{0.45\linewidth}
  \centering
  \includegraphics[width=1\linewidth]{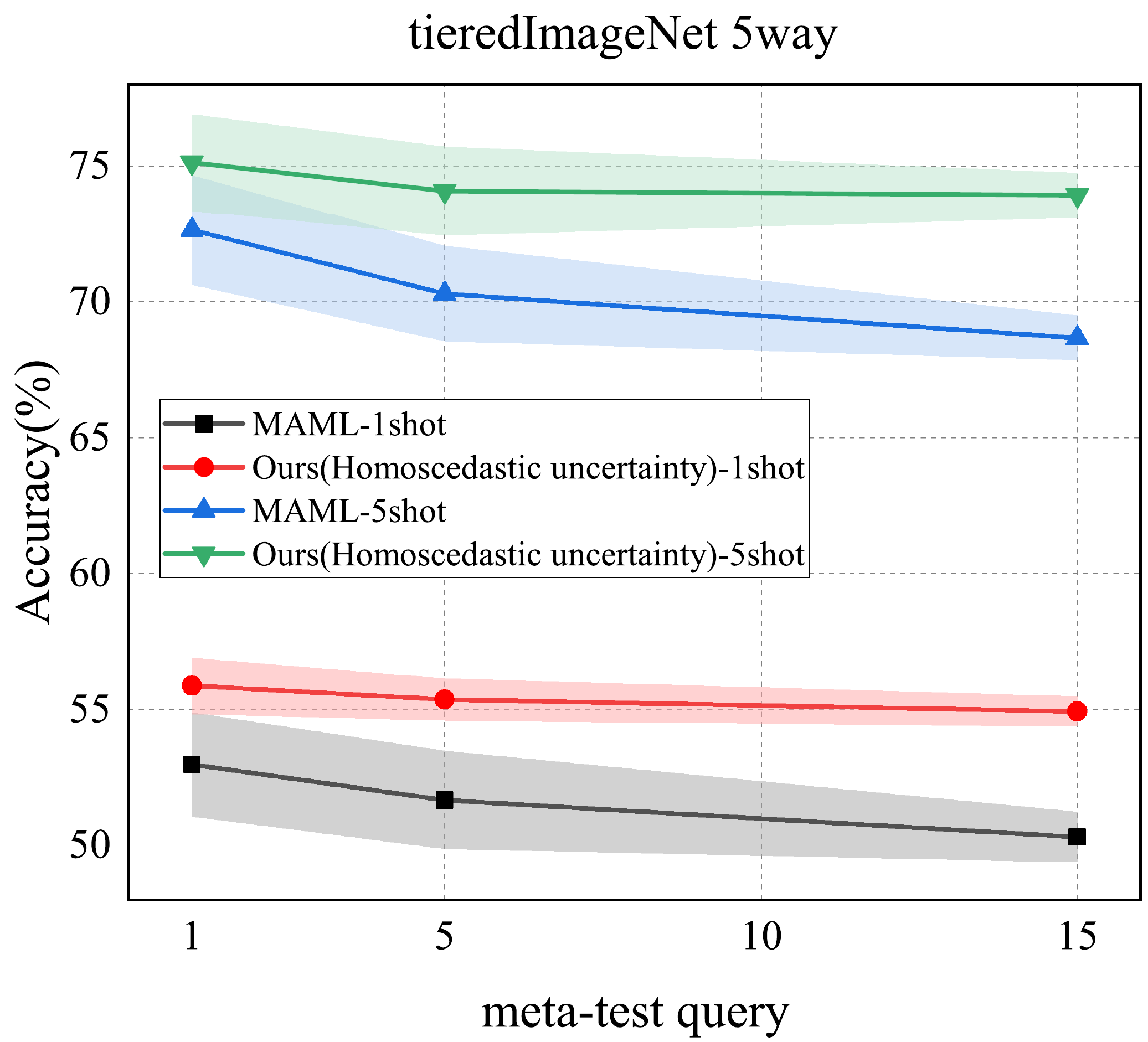}  
  \end{minipage} 
  }
  \vspace{-1em}
  \caption{Test accuracy ($\%$) of MAML and our (Homoscedastic uncertainty) method for different query sets of meta-test. The shaded area represents the 95$\%$ confidence interval. The performance of our method on both 1-shot and 5-shot training is extremely robust to the number of query sets.}
  \label{fig5.5}
\vspace{-1em}
\end{figure}

Based on the classification experiments in Section 5.2, we set different numbers of meta-test query sets (n$\_$query $\in \left\{1,5,15\right\}$) and conducted multiple sets of comparison experiments to observe the adaptability of our (homoscedastic uncertainty) method compared to MAML.

As shown in Fig.~\ref{fig5.5}, the accuracy of MAML decreased significantly from one query set to five query sets. In contrast, the performance of our (homoscedastic uncertainty) method on the 1-shot and 5-shot mechanisms of the miniImageNet and tieredImageNet datasets does not decrease significantly with an increase in the number of query sets. This makes our method more practical because it reduces the sensitivity to the number of query sets during meta-testing.

\section{Conclusions and Prospects}
In this study, we introduce a meta-loss function with homoscedastic uncertainty to improve the efficiency of meta-learning, which simultaneously maintains the generalization of knowledge through task-dependent uncertainty. Compared with several baselines, the experiments demonstrate the effectiveness and robustness of our algorithm for regression and few-shot classification problems.

Although our method has good performance, it has some limitations and interesting future directions. (1) In this paper, we introduce the loss function in multiple tasks, but do not consider how to avoid the problem of negative migration. (2) Another interesting direction is to add a linear classifier to MAML, a change that may greatly improve the ability of our model to generalize on larger and more complex datasets, increasing its use in increasingly complex task scenarios. We plan to continue to study these aspects in the future.

%
%

%
%
%
%

\end{document}